\documentclass[11pt]{article}
\usepackage[margin=1in]{geometry}
\usepackage{amsmath,amssymb,amsthm}
\usepackage{booktabs}
\usepackage{array}
\usepackage{tabularx}
\usepackage{float}
\usepackage{hyperref}
\usepackage{microtype}
\usepackage{bm}

\newcolumntype{Y}{>{\raggedright\arraybackslash}X}

\title{Physics-Aware Machine Learning for Seismic and Volcanic Signal Interpretation}
\author{William Thorossian}
\date{March 18, 2026}

\begin{document}
\maketitle

\begin{abstract}
Modern seismic and volcanic monitoring is increasingly shaped by continuous, multi-sensor observations and by the need to extract actionable information from nonstationary, noisy wavefields. In this context, machine learning has moved from a research curiosity to a practical ingredient of processing chains for detection, phase picking, classification, denoising, and anomaly tracking.

However, improved accuracy on a fixed dataset is not sufficient for operational use. Models must remain reliable under domain shift (new stations, changing noise, evolving volcanic activity), provide uncertainty that supports decision-making, and connect their outputs to physically meaningful constraints.

This paper surveys and organizes recent ML approaches for seismic and volcanic signal analysis, highlighting where classical signal processing provides indispensable inductive bias, how self-supervision and generative modeling can reduce dependence on labels, and which evaluation protocols best reflect transfer across regions. We conclude with open challenges for robust, interpretable, and maintainable AI-assisted monitoring.
\end{abstract}


\section{Introduction}
Seismic and volcanic monitoring increasingly relies on dense sensor deployments (broadband seismometers, infrasound arrays, GNSS, tiltmeters, and increasingly distributed acoustic sensing) and on continuous, long-duration archives. These data streams contain information spanning time scales from fractions of a second (high-frequency fracture and rockfall signals) to months (slow deformation and evolving tremor), but are also shaped by strong site effects, transient noise sources, and nonstationary propagation conditions.

Classical signal processing provides a rigorous toolbox for detection, characterization, and inversion, yet many operational tasks---event detection, phase picking, denoising, classification, and anomaly tracking---have become bottlenecks as data volume and heterogeneity grow. Machine learning (ML) and modern AI methods can complement physics-based approaches by learning robust representations directly from data, especially when coupled with explicit uncertainty estimates and physically meaningful constraints.

This paper proposes a practical technical framework, focusing on (i) signal processing foundations that remain essential, (ii) ML architectures and training strategies tailored to seismic and volcanic signals, (iii) ways to connect data-driven models with physical understanding, and (iv) evaluation protocols that encourage transfer across regions and instruments.

\section{Seismic and volcanic signals: what makes them difficult?}
Seismic and volcanic waveforms are challenging not only because of noise, but because the generating processes and the recording conditions evolve. In practice, the ``same'' physical phenomenon can look different across stations, seasons, and episodes of unrest, while different mechanisms can produce superficially similar signatures.

\subsection{Nonstationarity and mixed sources}
Volcanic environments are characterized by rapidly changing source processes (fracturing, fluid flow, conduit resonance, explosions) and by evolving paths as the near-surface structure changes with temperature, saturation, and damage. Even in tectonic settings, ambient noise and anthropogenic activity can vary strongly with time of day and season. A key implication is that models trained on one period may fail when monitoring conditions drift.

From a modeling perspective, nonstationarity appears at multiple levels: (i)\ baseline noise statistics drift, (ii)\ the target class distribution changes (e.g., an increase in tremor during unrest), and (iii)\ the mapping from waveform to label changes as the source mechanism itself evolves. Robust methods therefore need explicit monitoring of data quality and distribution shift rather than assuming fixed training and test conditions.

\subsection{Complex propagation and site effects}
Local 3D structure, topography, and strong heterogeneity near volcanoes can distort waveforms through scattering, attenuation, and mode conversions. The same event can therefore produce different apparent onset sharpness, frequency content, and polarization depending on station location and coupling.

These effects complicate both classical picking and ML classification: a model may latch onto station-specific spectral ``fingerprints'' rather than source properties. Practical mitigations include careful band selection, station-wise normalization, and evaluation that holds out entire stations or time periods to test whether the model has learned transferable features.

\subsection{Imbalanced labels and ambiguous taxonomies}
Rare but high-impact events (e.g., precursory signals to eruptions or large earthquakes) are sparsely labeled, while common classes (noise, small events) dominate. Moreover, volcanic event categories (LP, VT, hybrid, tremor, explosion) often overlap in the time--frequency domain and depend on local conventions, leading to label noise and limited comparability across observatories.

Ambiguity also arises because categories are often proxies for mechanisms rather than mechanisms themselves: ``LP'' may reflect a mixture of fluid-related resonance and path effects, and tremor can result from sustained degassing, rockfalls, or even cultural noise. This motivates hierarchical or probabilistic labeling schemes in which events receive distributions over types rather than single hard labels.

\subsection{Overlapping events, long durations, and context dependence}
Volcanic tremor and infrasound can persist for minutes to hours, while microseismicity may occur in swarms with overlapping signals. In such settings, the relevant ``event'' is not always a well-isolated window with clear boundaries.

Methods that treat analysis windows as independent samples can miss this context. Sequence models that incorporate temporal continuity (e.g., hidden-state formulations, temporal convolutions, or attention with long receptive fields) can improve consistency, but must be paired with constraints that prevent error propagation during extended runs.

\subsection{Array geometry, instrument response, and missing data}
Differences in sensor type, response, sampling rates, and installation conditions create domain shifts. Operational streams also contain gaps, clipped waveforms, timing errors, and changing channel availability, which must be handled explicitly rather than silently discarded.

In multi-station systems, geometry matters: azimuthal coverage affects location uncertainty and association performance, while aperture and spacing influence beamforming resolution. For ML, these factors suggest designing inputs that are robust to variable station sets (e.g., set-based encoders) and making missingness explicit (masking) rather than relying on ad hoc interpolation.

\subsection{Uncertainty in ``ground truth''}
Even when high-quality catalogs exist, picks and labels reflect analyst choices, network configuration, and evolving practices. For volcanic signals, the boundary between noise and low-amplitude events can be subjective, and retrospective relabeling is common during eruptive crises.

Consequently, evaluation should treat labels as imperfect observations. Reporting inter-analyst agreement (when available), using soft targets, and validating with injection tests (adding synthetic signals into real noise) can provide a more realistic view of model performance.
\section{Signal processing foundations (still the backbone)}
Signal processing remains the backbone of seismic and volcanic data analysis because it formalizes assumptions about wave propagation, noise, and sampling. In AI-enabled systems, these methods serve two complementary roles: they (i)\ provide high-SNR features and priors that reduce sample complexity, and (ii)\ define sanity checks that help detect out-of-distribution behavior.

\subsection{Preprocessing as model conditioning}
Standard steps include instrument response correction (when appropriate), consistent filtering, robust normalization, and careful windowing. For ML pipelines, these operations should be deterministic, logged, and reproducible so that model failures can be traced to data conditioning.

A recurring pitfall is to treat preprocessing as ``just data cleaning'' when it actually determines the invariances the model will learn. For instance, aggressive bandpass filtering can erase onset cues needed for picking, while per-window standardization can remove amplitude information that is diagnostically meaningful for explosions and tremor.

When possible, preprocessing should also carry metadata forward: quality flags, clipping indicators, and gap masks can be appended as auxiliary inputs so the model can learn to down-weight unreliable segments.

\subsection{Denoising, deglitching, and gap handling}
Continuous streams frequently contain impulsive artifacts (telemetry glitches, footsteps, electrical interference), long-term drifts, and missing segments. Classical approaches such as median-based despiking, robust detrending, spectral whitening, and notch filtering remain effective, but must be applied conservatively to avoid distorting true seismic phases.

For monitoring pipelines, it is useful to distinguish between (i)\ corrective operations that aim to reconstruct the underlying signal (e.g., interpolation across short gaps) and (ii)\ censoring operations that mark unusable data (e.g., masking saturated windows). The latter is often safer for downstream ML because it prevents models from learning from artifacts.

\subsection{Time--frequency representations}
Short-time Fourier transforms, wavelet transforms, and multitaper spectra remain central for tremor and resonance analysis. In ML settings, they can serve as (i) inputs to 2D convolutional models, (ii) targets for denoising or inpainting, or (iii) diagnostic views for interpretation.

Representation design should be treated as a modeling decision. Window length and overlap trade time resolution against frequency resolution; multitaper choices trade variance reduction against spectral leakage. For volcanic tremor, using multiple resolutions in parallel (e.g., a short-window and a long-window view) can capture both rapid modulation and stable resonance peaks.

\subsection{Polarization, coherence, and array processing}
Three-component polarization analysis, inter-station coherence, and array beamforming encode geometric constraints that are difficult to learn from a single station. Beamforming and FK analysis can separate coherent arrivals from incoherent noise and provide back-azimuth and slowness estimates that are valuable both as features and as validation signals.

A practical hybrid approach is to compute beamformed stacks or slowness traces and feed them to ML models alongside raw waveforms. This retains physical interpretability while allowing the learned model to cope with nonidealities (e.g., partial station outages).

\subsection{Detection, picking, and association as a coupled problem}
Matched filtering, STA/LTA variants, and array detectors provide strong priors for detecting low-SNR signals. However, in dense sequences and swarms, detection, phase picking, and event association are coupled: small pick biases can fragment an event into multiple hypotheses, while missed detections can break associations.

One strategy is a two-stage pipeline: a conservative detector proposes candidates, then a learned model refines picks and class labels, and finally a physics-based associator enforces travel-time consistency (e.g., PhaseLink and GaMMA) \cite{ross2019phaselink,zhu2022gamma}. An alternative is joint optimization in which association constraints feed back into picking (e.g., penalizing picks that imply implausible velocity structures).

\subsection{Feature engineering as inductive bias (not a relic)}
Engineered features such as envelope statistics, spectral ratios, kurtosis, and polarization attributes remain valuable, especially in low-label regimes. Rather than competing with deep learning, these features can be used to (i)\ initialize or regularize learned representations, (ii)\ provide interpretable baselines, and (iii)\ serve as ``concepts'' for post hoc explanations.

\section{Machine learning methods tailored to seismology}
\subsection{Supervised learning for picks, detection, and classification}
For phase picking and event detection, modern architectures include 1D convolutional networks, temporal convolutional networks, and attention-based models \cite{zhu2018phasenet,mousavi2020eqtransformer}. Key design choices are the receptive field (to capture emergent onsets), the treatment of multi-component inputs, and calibration of predicted probabilities so that thresholds correspond to interpretable false-alarm rates.

For volcanic classification, hierarchical labeling can reduce ambiguity (e.g., first distinguish event vs.\ noise, then impulsive vs.\ tremor-like, then subtypes) \cite{zhong2024volcanophase,tan2024pavlof,fee2025voissnet}. Multi-task learning can further stabilize training by predicting auxiliary quantities such as SNR, dominant frequency, or polarization attributes.

\subsection{Self-supervised and weakly supervised learning}
Because labels are scarce and inconsistent, self-supervised learning is particularly attractive. Pretext tasks include masked-sample reconstruction, contrastive learning across channels or nearby stations, and time-shift prediction. These approaches can learn representations that transfer to new regions with minimal fine-tuning, provided that augmentation strategies respect the physics (e.g., polarity flips and time reversal may not be valid for all tasks). Recent large-scale pretraining studies suggest that foundation-model backbones can provide strong initialization across multiple downstream tasks (detection, picking, and location) \cite{liu2024seislm,saad2026utrans}.

Weak supervision leverages catalog heuristics (e.g., simple triggers, analyst flags, eruption intervals) to create noisy labels at scale. Robust loss functions and label-noise models can prevent the network from overfitting these heuristics.

\subsection{Generative models for denoising and simulation}
Denoising autoencoders and diffusion models can remove transient noise, fill gaps, and standardize waveform quality \cite{zhu2019deepdenoiser}. A crucial requirement is to avoid hallucinating physically implausible phases or spectral content. Practically, this means (i) conditioning on station metadata and band limits, (ii) validating on synthetic injections, and (iii) reporting uncertainty bands rather than point estimates.

Generative models can also serve as fast surrogates for forward modeling in limited contexts (e.g., approximating site-specific impulse responses), but they should not replace physical simulation when extrapolating beyond the training distribution.

\section{Bridging data-driven models and physical understanding}
A central challenge in AI-enabled monitoring is to prevent models from becoming ``black boxes'' that perform well only within a narrow training distribution. Bridging data-driven models with physical understanding can be viewed as adding structure to an otherwise flexible function approximator.

Let $\mathcal{D}=\{(\mathbf{x}_i,\mathbf{y}_i)\}_{i=1}^{N}$ denote a labeled dataset and let $f_{\theta}$ be a predictor. A generic learning objective can be written as
\begin{equation}
  \min_{\theta}\; \frac{1}{N}\sum_{i=1}^{N} \ell\bigl(f_{\theta}(\mathbf{x}_i),\mathbf{y}_i\bigr) + \lambda\,\Omega(\theta),
\end{equation}
where $\Omega$ encodes prior knowledge or desired behavior. In this section, we describe practical choices for $\Omega$ and for post hoc diagnostics.

\subsection{Physics-informed constraints}
Physical knowledge can enter the learning problem through constrained outputs (e.g., enforcing nonnegative duration, bounded velocities), through differentiable forward operators (e.g., travel-time models), or through regularization terms that penalize violations of known relationships.

\paragraph{Consistency constraints for picks and locations.}
Suppose a model predicts phase arrival times $\hat{t}_{s,p}$ for station $s$ and phase $p\in\{P,S\}$. For a candidate hypocenter $\mathbf{h}$ and origin time $t_0$, a travel-time model $\tau_p(\mathbf{h},s)$ implies expected arrivals $t_0+\tau_p(\mathbf{h},s)$. A physically motivated penalty is
\begin{equation}
  \mathcal{L}_{\mathrm{tt}} = \sum_{s\in\mathcal{S}}\sum_{p\in\{P,S\}} \rho\bigl(\hat{t}_{s,p} - (t_0+\tau_p(\mathbf{h},s))\bigr),
\end{equation}
where $\rho$ is a robust loss (e.g., Huber) and $\mathcal{S}$ is the set of available stations. Minimizing $\mathcal{L}_{\mathrm{tt}}$ can be done as a post-processing step (quality control) or integrated into training when $\tau_p$ is differentiable.

\paragraph{Amplitude and attenuation structure.}
For some tasks, relative amplitudes carry physical meaning. A simple attenuation proxy is
\begin{equation}
  \log A_s \approx \log A_0 - \kappa\,r_s,
\end{equation}
where $A_s$ is an observed amplitude measure at station $s$, $r_s$ is source--receiver distance, and $\kappa$ summarizes geometrical spreading and attenuation. Models that predict source strength can be regularized to respect monotonic decay with distance (within uncertainty), rather than exploiting station idiosyncrasies.

\paragraph{Differentiable operators and hybrid losses.}
When parts of the forward problem are differentiable, one can train models to predict latent physical parameters and compare in observation space. For example, predicting a source spectrum $S(f)$ and comparing to observed spectra $X_s(f)$ via
\begin{equation}
  \mathcal{L}_{\mathrm{spec}}=\sum_{s\in\mathcal{S}} \int \bigl\lvert X_s(f) - G_s(f)\,S(f)\bigr\rvert^2\,df,
\end{equation}
with station-dependent transfer functions $G_s(f)$, constrains the model to produce physically plausible outputs.

\subsection{Uncertainty and decision support}
Operational monitoring needs calibrated uncertainty: decision-makers require not just a label, but a defensible estimate of error.

\paragraph{Predictive and epistemic uncertainty.}
With an ensemble $\{f_{\theta^{(m)}}\}_{m=1}^{M}$, the predictive mean for a scalar output is
\begin{equation}
  \bar{y}(\mathbf{x})=\frac{1}{M}\sum_{m=1}^{M} f_{\theta^{(m)}}(\mathbf{x}),
\end{equation}
and the ensemble variance
\begin{equation}
  \mathrm{Var}(\mathbf{x})=\frac{1}{M}\sum_{m=1}^{M} \bigl(f_{\theta^{(m)}}(\mathbf{x})-\bar{y}(\mathbf{x})\bigr)^2
\end{equation}
can be used as a proxy for epistemic uncertainty. In practice, large variance can trigger analyst review or fallback to conservative signal-processing detectors.

\paragraph{Calibration.}
For probabilistic classifiers, calibration can be assessed via reliability diagrams or scalar metrics such as the expected calibration error (ECE). If $\hat{p}$ are predicted confidences and $\hat{y}$ are predicted labels, ECE partitions predictions into bins $B_b$ and measures
\begin{equation}
  \mathrm{ECE}=\sum_{b} \frac{|B_b|}{n}\,\bigl\lvert \mathrm{acc}(B_b)-\mathrm{conf}(B_b)\bigr\rvert.
\end{equation}
Beyond point calibration, conformal prediction can provide distribution-free prediction sets with explicit coverage guarantees, an attractive property for risk-aware monitoring under shift \cite{angelopoulos2021conformal,singh2024conformaleo,myren2025seismicuq}. Well-calibrated models support meaningful alert thresholds (e.g., ``issue an alert if $p>0.99$'') \cite{guo2017calibration}.

\paragraph{Decision-aware objectives.}
When the cost of false alarms and missed events is asymmetric, training can incorporate a cost matrix $C(k,\hat{k})$ through an expected risk
\begin{equation}
  \mathcal{R}(\theta)=\frac{1}{N}\sum_{i=1}^{N}\sum_{k=1}^{K} p_{ik}\,C\bigl(k,\hat{k}_i\bigr),
\end{equation}
or can adjust thresholds post hoc to match operational constraints.

\subsection{Interpretability beyond saliency maps}
For seismic applications, interpretability should connect to signal characteristics: which time--frequency bands, phases, or polarization features drive a decision?

\paragraph{Perturbation and counterfactual tests.}
Given an input $\mathbf{x}$ and a perturbation operator $T_{\delta}$ (e.g., notch out a frequency band, mask a time interval, rotate components), one can measure sensitivity
\begin{equation}
  \Delta(\delta)=\ell\bigl(f_{\theta}(T_{\delta}(\mathbf{x})),\hat{y}\bigr)-\ell\bigl(f_{\theta}(\mathbf{x}),\hat{y}\bigr),
\end{equation}
to identify which aspects of the signal are truly used by the model.

\paragraph{Concept-based explanations.}
Let $c(\mathbf{x})$ be a physically meaningful ``concept'' such as dominant frequency, polarization rectilinearity, or envelope kurtosis. One can quantify alignment between model representations $\mathbf{z}=h_{\theta}(\mathbf{x})$ and concepts by learning a linear probe $a$ such that $a^\top\mathbf{z}\approx c(\mathbf{x})$, providing an interpretable bridge between deep features and classical descriptors.

\paragraph{Sanity checks and stress tests.}
Interpretability should be paired with stress tests: (i)\ swap station identities, (ii)\ evaluate on periods with different noise regimes, and (iii)\ inject synthetic events of controlled amplitude and frequency. Models that are stable under these tests are more likely to generalize.

\section{Generalization and transfer across regions}
Generalization is often the limiting factor for deploying ML in monitoring: errors are rarely caused by the architecture alone, but by mismatch between training conditions and operational reality. In seismology, this mismatch is not exceptional---it is the norm.

\subsection{Domain shift as the default}
Differences in geology, site response, instrument type, and noise environment mean that cross-region generalization is usually harder than in-domain accuracy. Therefore, evaluation should include explicit out-of-domain tests, not only random train/test splits from the same stations and time periods.

\paragraph{A simple decomposition of shift.}
Let $P_{\mathrm{tr}}(\mathbf{x},y)$ and $P_{\mathrm{te}}(\mathbf{x},y)$ be the joint distributions at training and deployment. Generalization failures arise when $P_{\mathrm{tr}}\neq P_{\mathrm{te}}$. Common cases include:
\begin{itemize}
  \item \textbf{Covariate shift:} $P_{\mathrm{tr}}(\mathbf{x})\neq P_{\mathrm{te}}(\mathbf{x})$ but $P(y\mid\mathbf{x})$ is stable (e.g., a new noise regime).
  \item \textbf{Prior shift:} class proportions change, $P_{\mathrm{tr}}(y)\neq P_{\mathrm{te}}(y)$ (e.g., swarm onset increases event rate).
  \item \textbf{Concept drift:} the conditional changes, $P_{\mathrm{tr}}(y\mid\mathbf{x})\neq P_{\mathrm{te}}(y\mid\mathbf{x})$ (e.g., evolving tremor mechanism).
\end{itemize}
In practice, more than one type occurs simultaneously.

\paragraph{Evaluation protocols that reflect operations.}
To avoid optimistic estimates, splits should respect station, time, and region structure:
\begin{itemize}
  \item \textbf{Leave-one-station-out:} train on a subset of stations and test on unseen stations.
  \item \textbf{Time-blocked splits:} train on early periods and test on later periods to expose drift.
  \item \textbf{Leave-one-region-out:} test transfer across volcanoes or tectonic provinces.
\end{itemize}
For each protocol, reporting performance as a function of SNR and distance (when available) helps identify whether failures occur for weak signals or for specific domains.

\subsection{Strategies for transfer}
Effective strategies include normalization with station metadata, transfer learning/domain adaptation, and fine-tuning with small curated sets \cite{chai2020transfer,zhu2023phasenetdas,ganin2016dann,munchmeyer2022whichpicker}.

\paragraph{Metadata-aware normalization.}
Let $\mathbf{u}_s$ denote station metadata (e.g., sensor type, site class, sampling rate). A model can be conditioned as $f_{\theta}(\mathbf{x},\mathbf{u}_s)$ so that station-specific effects are represented explicitly rather than implicitly. In practice, this can be implemented by concatenating embeddings of metadata to intermediate feature maps.

\paragraph{Importance weighting for covariate shift.}
Under covariate shift, one can weight training examples by $w(\mathbf{x})=P_{\mathrm{te}}(\mathbf{x})/P_{\mathrm{tr}}(\mathbf{x})$ and minimize
\begin{equation}
  \min_{\theta}\; \frac{1}{N}\sum_{i=1}^{N} w(\mathbf{x}_i)\,\ell\bigl(f_{\theta}(\mathbf{x}_i),y_i\bigr).
\end{equation}
Although $w$ is rarely known exactly, it can be approximated by density-ratio estimation or by training a discriminator to separate training from target data.

\paragraph{Domain-adversarial learning.}
Let $h_{\theta}$ be a feature extractor, $g$ a label head, and $d_{\phi}$ a domain classifier predicting domain $s\in\{\text{source},\text{target}\}$. Domain-adversarial training seeks features that are predictive of $y$ but uninformative about $s$ \cite{ganin2016dann}:
\begin{equation}
  \min_{\theta,g}\max_{\phi}\; \mathcal{L}_{y}\bigl(g(h_{\theta}(\mathbf{x})),y\bigr)-\beta\,\mathcal{L}_{s}\bigl(d_{\phi}(h_{\theta}(\mathbf{x})),s\bigr).
\end{equation}
This approach can reduce sensitivity to station-specific spectra, but may also remove genuinely informative physical differences; it should therefore be validated with out-of-domain tests.

\paragraph{Fine-tuning and continual adaptation.}
For observatories, a practical workflow is ``train globally, adapt locally'': a shared base model is trained on multi-region data, then local calibration layers are updated as the network evolves. To prevent catastrophic forgetting, one can fine-tune with regularization toward the base parameters $\theta_0$, for example via
\begin{equation}
  \mathcal{L}_{\mathrm{ft}}(\theta)=\mathcal{L}_{\mathrm{target}}(\theta)+\gamma\,\lVert \theta-\theta_0\rVert_2^2.
\end{equation}

\paragraph{Open-set recognition for ``unknown'' signals.}
Deployment inevitably encounters signals not represented in the training set. A practical safeguard is to treat unfamiliar inputs as ``unknown'' when confidence is low or when uncertainty is high (e.g., using ensemble variance or a distance-to-prototype score in embedding space), routing these cases for analyst review.

\section{Comparative synthesis and mini-validation}
To make this survey more actionable for method selection, we include a compact comparison framework and a minimal validation protocol that can be reused across observatories.

\subsection{Systematic comparison of representative approaches}
Table~\ref{tab:method_comparison} summarizes representative families of methods along dimensions that matter operationally: strengths, limits, robustness to domain shift, interpretability, and deployment readiness. Representative references include PhaseNet/EQTransformer \cite{zhu2018phasenet,mousavi2020eqtransformer}, PhaseLink/GaMMA \cite{ross2019phaselink,zhu2022gamma,puentehuerta2025associators}, transfer/adaptation studies \cite{chai2020transfer,zhu2023phasenetdas,ganin2016dann}, uncertainty-aware approaches \cite{guo2017calibration,angelopoulos2021conformal,singh2024conformaleo,myren2025seismicuq}, foundation pretraining \cite{liu2024seislm,saad2026utrans}, and volcanic multi-site models \cite{tan2024pavlof,fee2025voissnet}.

\begin{table}[htbp]
\centering
\footnotesize
\renewcommand{\arraystretch}{1.15}
\setlength{\tabcolsep}{3pt}
\begin{tabularx}{\textwidth}{>{\raggedright\arraybackslash}p{1.8cm}>{\raggedright\arraybackslash}p{1.5cm}>{\raggedright\arraybackslash}p{1.75cm}YY>{\centering\arraybackslash}p{1.35cm}>{\centering\arraybackslash}p{1.35cm}>{\centering\arraybackslash}p{1.5cm}}
\toprule
Task & Input & Model & Strengths & Limits & \shortstack{Shift\\robust.} & \shortstack{Interpret-\\ability} & \shortstack{Ops\\readiness} \\
\midrule
Detection \& picking & 3C windows & CNN / Transformer & High recall with mature tooling & Station bias without strict split design & Medium & Medium & High \\
Association & Picks + metadata & Sequence / BGMM & Event-level consistency in swarms & Upstream pick errors propagate & \shortstack{Medium--\\High} & High & High \\
Transfer / adaptation & Source + target + metadata & Transfer + DANN & Label-efficient adaptation to new regions & Negative transfer; tuning overhead & High$^{\dagger}$ & Medium & Medium \\
Uncertainty-aware classification & Scores + calibration split & Ensemble + conformal & Risk-aware thresholding and coverage control & Extra computation under severe shift & \shortstack{Medium--\\High} & High & \shortstack{Medium--\\High} \\
Foundation representations & Unlabeled waveform archives & SeisLM-like encoders & Strong few-label transfer across tasks & Expensive pretraining; limited common benchmarks & \shortstack{Promis-\\ing} & \shortstack{Low--\\Medium} & Medium \\
Volcanic multi-site classification & Seismic / infrasound spectrograms & VOISS-Net-like models & Broad cross-volcano detection coverage & Confusions on unseen transient classes & Medium & Medium & \shortstack{Medium--\\High} \\
\bottomrule
\end{tabularx}
\caption{Comparative synthesis of representative ML approaches for seismic and volcanic monitoring. Ratings are qualitative and intended as survey-level guidance rather than leaderboard claims. $^{\dagger}$High when validated with strict out-of-domain protocols.}
\label{tab:method_comparison}
\end{table}

\subsection{Mini case study: split protocol, calibration, and physics constraints}
Even in a survey paper, a compact conceptual validation can sharpen scientific claims. A minimal protocol can be run with one architecture and fixed training budget under four settings: (i) random train/test split, (ii) leave-one-station-out (LOSO), (iii) LOSO + post hoc confidence calibration, and (iv) LOSO + calibration + a physics-aware hybrid objective.

For the hybrid setting, one can combine task and travel-time consistency losses as
\begin{equation}
  \mathcal{L}_{\mathrm{hyb}} = \mathcal{L}_{\mathrm{task}} + \lambda_{\mathrm{tt}}\,\mathcal{L}_{\mathrm{tt}}.
\end{equation}

A concise report should include: (a) random-vs.-LOSO performance gap (to expose leakage), (b) confidence quality before/after calibration (ECE and reliability curves), and (c) physical consistency metrics (e.g., travel-time residual tails and outlier rates). A convincing pattern is typically: random split appears optimistic, LOSO reveals failure under shift, calibration improves decision reliability with minimal loss in discrimination, and hybrid loss improves physical consistency in difficult cases \cite{munchmeyer2022whichpicker,guo2017calibration,angelopoulos2021conformal}.

\section{A reference workflow for reproducible studies}
Reproducibility in AI-driven seismology is not only about sharing code; it requires making each processing and modeling decision traceable to specific data versions, parameters, and evaluation protocols \cite{mousavi2019stead,woollam2022seisbench,michelini2021instance}. We propose a workflow organized into four layers, each producing explicit artifacts that can be archived and audited.

\begin{figure}[H]
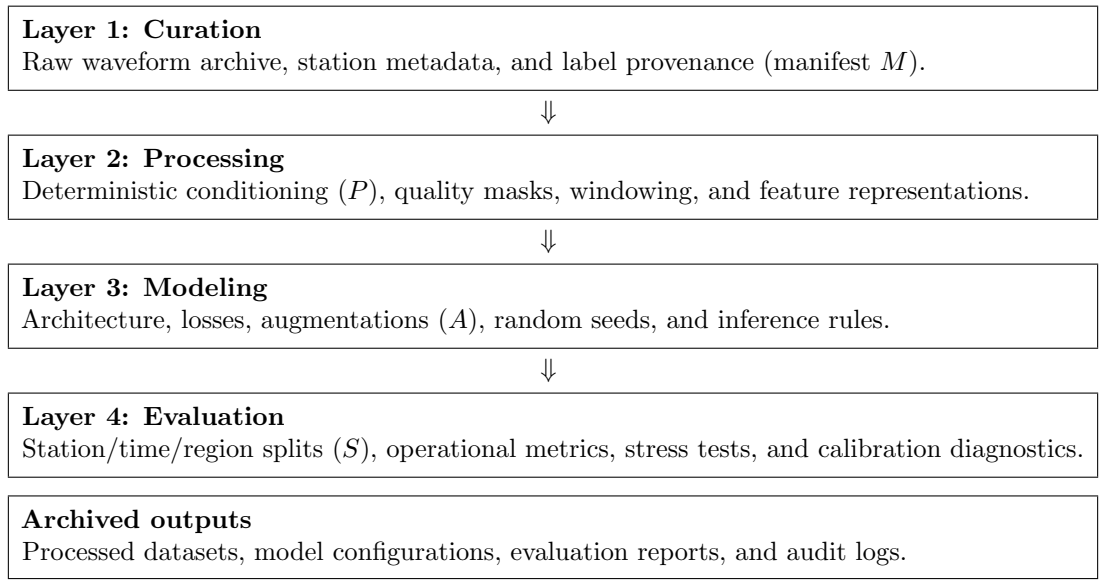

\centering
\small
\begin{minipage}{0.95\linewidth}
\centering
\setlength{\fboxsep}{5pt}
\fbox{\parbox{0.90\linewidth}{\textbf{Layer 1: Curation}\\Raw waveform archive, station metadata, and label provenance (manifest $M$).}}

\vspace{1mm}
$\Downarrow$
\vspace{1mm}

\fbox{\parbox{0.90\linewidth}{\textbf{Layer 2: Processing}\\Deterministic conditioning ($P$), quality masks, windowing, and feature representations.}}

\vspace{1mm}
$\Downarrow$
\vspace{1mm}

\fbox{\parbox{0.90\linewidth}{\textbf{Layer 3: Modeling}\\Architecture, losses, augmentations ($A$), random seeds, and inference rules.}}

\vspace{1mm}
$\Downarrow$
\vspace{1mm}

\fbox{\parbox{0.90\linewidth}{\textbf{Layer 4: Evaluation}\\Station/time/region splits ($S$), operational metrics, stress tests, and calibration diagnostics.}}

\vspace{2mm}
\fbox{\parbox{0.90\linewidth}{\textbf{Archived outputs}\\Processed datasets, model configurations, evaluation reports, and audit logs.}}
\end{minipage}
\caption{Conceptual workflow for reproducible AI-assisted seismic and volcanic monitoring. Each layer has explicit inputs, transformations, and archived outputs.}
\label{fig:workflow_concept}
\end{figure}

Figure~\ref{fig:workflow_concept} summarizes this workflow as a practical pipeline from curation to operational evaluation.

\subsection{Layer 1: data curation}
\textbf{Goal:} define what the dataset is.

At minimum, curation should record: station/channel lists, sensor types and responses, sampling rates, timing corrections, gap statistics, and any deglitching or clipping detection. For labels, it is essential to document provenance: catalog version, analyst procedure, relabeling policies, and known ambiguities.

A practical approach is to define a dataset manifest $M$ containing hashes of raw files and metadata, so that the curated dataset is a deterministic function of the archive. Conceptually, one can write
\begin{equation}
  (\mathcal{X},\mathcal{Y}) = \Phi_{\mathrm{cur}}(\text{archive}; M),
\end{equation}
where $\Phi_{\mathrm{cur}}$ is the curation pipeline.

\subsection{Layer 2: processing}
\textbf{Goal:} make waveform conditioning and representations reproducible.

Processing choices (filter corners, response removal, whitening, window length, overlap, spectrogram parameters) should be stored in a processing configuration $P$. The processed dataset becomes
\begin{equation}
  \tilde{\mathcal{X}} = \Phi_{\mathrm{proc}}(\mathcal{X}; P).
\end{equation}
When possible, store intermediate products (e.g., spectrograms or quality masks) or ensure they can be regenerated exactly from $M$ and $P$.

\subsection{Layer 3: modeling}
\textbf{Goal:} specify the training objective and all sources of randomness.

A model is defined not only by its architecture, but also by its loss functions, optimizer, learning-rate schedule, augmentations, batch construction rules, and random seeds. Let $A$ denote the set of augmentation parameters and $H$ denote hyperparameters (including $\lambda$ weights for multi-task losses). Training can be summarized as
\begin{equation}
  \theta^{\star} = \arg\min_{\theta}\; \mathbb{E}_{(\mathbf{x},\mathbf{y})\sim (\tilde{\mathcal{X}},\mathcal{Y})}\,\mathbb{E}_{\xi\sim A}\,\ell\bigl(f_{\theta}(T_{\xi}(\mathbf{x})),\mathbf{y}; H\bigr),
\end{equation}
where $T_{\xi}$ is an augmentation transform.

For monitoring, it is also important to record the inference configuration: windowing strategy, overlap, thresholds, and post-processing (association, smoothing, alert rules). These choices determine latency and false-alarm behavior.

\subsection{Layer 4: evaluation}
\textbf{Goal:} measure performance in a way that matches operations and transfer.

Evaluation should include in-domain and out-of-domain tests and a clear split specification $S$ (station/time/region blocks). Report not only aggregate metrics, but stratified results by SNR, distance, station, and time period.

For detection tasks, a useful operational metric is probability of detection at fixed false-alarm rate. If $\mathrm{FAR}(\tau)$ and $\mathrm{POD}(\tau)$ denote false-alarm rate and detection probability at threshold $\tau$, then one reports $\mathrm{POD}(\tau^{\star})$ where $\tau^{\star}$ satisfies $\mathrm{FAR}(\tau^{\star})=\mathrm{FAR}_0$.

For phase picking, summarize residual distributions (median, MAD) and tail probabilities (e.g., $P(|\Delta t|>0.5\,\mathrm{s})$). For volcanic classification, complement confusion matrices with station-wise performance and with drift analyses during changing activity.

\subsection{Recommended artifacts}
To make the workflow actionable, we recommend archiving:
\begin{itemize}
  \item Dataset manifest $M$ and label provenance notes.
  \item Processing configuration $P$ and scripts/pipelines $\Phi_{\mathrm{cur}},\Phi_{\mathrm{proc}}$.
  \item Model specification (architecture + losses), hyperparameters $H$, augmentation set $A$, and seeds.
  \item Split specification $S$, evaluation scripts, and per-domain metric tables.
\end{itemize}

\section{Open challenges and research directions}
Despite rapid progress, several challenges remain central before AI can be considered a mature, dependable component of seismic and volcanic monitoring.

\subsection{Robustness under nonideal telemetry and extremes}
Robustness is not only performance under additive noise; it includes resilience to clipping, timing errors, dropped packets, variable sampling rates, and changing channel availability.

Key open problems include: (i)\ detecting when a model is operating outside its validated regime; (ii)\ designing models that degrade gracefully rather than catastrophically; and (iii)\ building standardized stress tests that reflect real failure modes. For example, one can define corruption operators $T_{\delta}$ (gap insertion, clipping, notch interference) and evaluate worst-case performance over a corruption set $\Delta$:
\begin{equation}
  \mathrm{Risk}_{\Delta}(f)=\sup_{\delta\in\Delta}\;\mathbb{E}\bigl[\ell\bigl(f(T_{\delta}(\mathbf{x})),y\bigr)\bigr].
\end{equation}

\subsection{Labeling, taxonomies, and benchmark datasets}
Progress is limited by inconsistent labels across networks and by the absence of widely used, cross-observatory benchmarks for volcanic signals.

Two priorities are (i)\ developing interoperable taxonomies (possibly hierarchical and probabilistic) and (ii)\ curating benchmark datasets with explicit station/time/region splits. Recent work has started to move in this direction through cross-dataset picker benchmarks and volcanic multi-site evaluations \cite{munchmeyer2022whichpicker,zhong2024volcanophase,tan2024pavlof,fee2025voissnet}, but broader shared benchmarks (including association components) remain needed for robust comparison \cite{puentehuerta2025associators}.

\subsection{From pattern recognition to mechanism}
Many current systems perform pattern recognition (``what class is this?'') but do not support mechanistic inference (``what process is changing?''). Bridging this gap requires models that can test hypotheses and support causal reasoning.

Promising directions include: (i)\ hybrid models that expose latent variables interpretable as physical parameters (e.g., resonance frequency, attenuation proxies); (ii)\ counterfactual simulation to test sensitivity to hypothesized changes (e.g., shifting a resonance peak); and (iii)\ integrating ML outputs into inverse problems where uncertainty is propagated to physical quantities.

\subsection{Real-time constraints and edge deployment}
Operational systems face strict latency, reliability, and compute constraints. Models must run continuously, handle data gaps, and provide bounded inference time.

Research questions include compressing models without losing calibration, designing streaming architectures with state, and quantifying the trade-off between window length (better context) and detection latency (faster response). In some settings, partial edge deployment can reduce bandwidth requirements, but only if on-device decisions are conservative and uncertainty-aware.

\subsection{Human--AI interaction and responsible use}
Monitoring centers are socio-technical systems: humans remain responsible for decisions, and ML outputs influence attention and interpretation.

Responsible use requires (i)\ calibrated uncertainty and clear communication of failure modes; (ii)\ interfaces that prevent automation bias (e.g., showing alternative explanations and confidence intervals); and (iii)\ logging and audit trails so that alerts can be reviewed retrospectively.

\subsection{Reproducibility and long-term maintainability}
Finally, operational value depends on maintainability: models must be retrained as networks change, and performance must be monitored over time. This motivates ``model lifecycle'' practices such as periodic re-validation, drift detection, and versioning of both data and models.

\subsection{Challenge--strategy matrix for operations}
Table~\ref{tab:challenge_strategy} maps recurring operational challenges to practical mitigation strategies and monitoring metrics, so that model governance is explicit rather than ad hoc.

\begin{table}[htbp]
\centering
\footnotesize
\setlength{\tabcolsep}{3pt}
\begin{tabular}{p{2.9cm}p{3.9cm}p{4.8cm}p{3.4cm}}
\hline
Challenge & Operational failure mode & Recommended strategies & Monitoring metrics \\
\hline
Domain shift (station/time/region) & Recall drop on unseen stations or new noise regimes & Leave-one-station-out and time-blocked validation; metadata-aware normalization; targeted local fine-tuning & POD at fixed FAR by station, drift indices \\
Uncalibrated confidence & Overconfident false alerts or missed events in rare regimes & Temperature scaling/ensembles; conformal prediction sets; ``unknown'' rejection policy & ECE, reliability slope, conformal coverage/set size \\
Noisy labels and ambiguous taxonomy & Brittle boundaries and poor cross-observatory transfer & Soft/hierarchical labels; cross-analyst audits; weak-label robust losses & Inter-annotator agreement, class-wise calibration \\
Telemetry artifacts and missing data & False triggers from glitches, silent failures on clipped/gappy streams & Explicit masking; corruption stress tests; conservative censoring over aggressive interpolation & Corruption risk curves, missing-channel robustness \\
Weak physical consistency & Plausible class scores but implausible picks/locations & Hybrid losses with travel-time constraints; post hoc association consistency checks & Travel-time residual tails, association consistency rate \\
Lifecycle and maintainability & KPI decay after network/configuration updates & Versioned data/models; periodic revalidation; rollback-ready deployment policy & Rolling OOD score, monthly KPI deltas \\
\hline
\end{tabular}
\caption{Challenges-versus-strategies summary for operational ML in seismic and volcanic monitoring.}
\label{tab:challenge_strategy}
\end{table}

\section{Conclusions}
AI-driven methods are becoming a practical component of seismic and volcanic monitoring, but their value depends on careful integration with signal processing, transparent evaluation under domain shift, and explicit links to physical interpretation.

A recurring theme throughout this paper is that the goal is not to replace physical reasoning, but to strengthen it: signal processing encodes decades of insight about sampling, propagation, and noise, while ML provides flexible representations and scalable automation. The most effective systems are therefore hybrid by design, combining physics-aware conditioning, constrained inference, and data-driven feature learning.

From an operational perspective, three practices appear essential for trustworthy deployment:
\begin{enumerate}
  \item \textbf{Explicit uncertainty:} models should report calibrated confidence and provide a mechanism to flag unfamiliar conditions (high epistemic uncertainty).
  \item \textbf{Evaluation under shift:} performance should be assessed with station-, time-, and region-held-out protocols, with stress tests reflecting telemetry artifacts and extreme noise.
  \item \textbf{Lifecycle management:} models and datasets should be versioned, monitored for drift, and periodically revalidated as networks and volcanic states evolve.
\end{enumerate}

For future survey updates, keeping a living comparison table and a fixed mini-validation protocol can make robustness claims auditable across model generations.

Finally, the most scientifically valuable direction is to move beyond classification toward inference about processes. This requires models that are interpretable in terms of time--frequency structure, polarization, and propagation constraints, and that can propagate uncertainty into downstream inverse problems. Achieving this will require collaboration between observatories, method developers, and domain scientists, as well as shared benchmarks and transparent reporting standards.

\section*{Acknowledgements and Disclosure of Funding}
This manuscript received no specific external funding. The author declares no competing interests.


\begin{thebibliography}{99}

\bibitem{angelopoulos2021conformal}
Anastasios N. Angelopoulos and Stephen Bates.
\newblock A gentle introduction to conformal prediction and distribution-free uncertainty quantification.
\newblock arXiv preprint arXiv:2107.07511, 2021.

\bibitem{chai2020transfer}
Chengping Chai, Monica Maceira, Hector J. Santos-Villalobos, Singanallur V. Venkatakrishnan, Martin Schoenball, Weiqiang Zhu, Gregory C. Beroza, and Clifford Thurber.
\newblock Using a deep neural network and transfer learning to bridge scales for seismic phase picking.
\newblock Geophysical Research Letters, 47(16):e2020GL088651, 2020.
\newblock doi: 10.1029/2020GL088651.

\bibitem{fee2025voissnet}
David Fee, Zhongwen Zhan, James Dixon, Luke Toney, Tianlin Liu, Jannes M{\"u}nchmeyer, Chris Marone, Christina Neal, Alex Witsil, Julian Dixon, and Cheryl Searcy.
\newblock A generalized deep learning model for volcanic seismicity detection and classification.
\newblock Volcanica, 8(1):305--323, 2025.
\newblock doi: 10.30909/vol/rjss1878.

\bibitem{ganin2016dann}
Yaroslav Ganin, Evgeniya Ustinova, Hana Ajakan, Pascal Germain, Hugo Larochelle, Fran{\c{c}}ois Laviolette, Mario Marchand, and Victor Lempitsky.
\newblock Domain-adversarial training of neural networks.
\newblock Journal of Machine Learning Research, 17(59):1--35, 2016.

\bibitem{guo2017calibration}
Chuan Guo, Geoff Pleiss, Yu Sun, and Kilian Q. Weinberger.
\newblock On calibration of modern neural networks.
\newblock In Proceedings of the 34th International Conference on Machine Learning (ICML), 2017.

\bibitem{liu2024seislm}
Tianlin Liu, Jannes M{\"u}nchmeyer, Laura Laurenti, Chris Marone, Maarten V. de Hoop, and Ivan Dokmani{\'c}.
\newblock SeisLM: A foundation model for seismic waveforms.
\newblock arXiv preprint arXiv:2410.15765, 2024.
\newblock doi: 10.48550/arXiv.2410.15765.

\bibitem{michelini2021instance}
Alberto Michelini et al.
\newblock INSTANCE: The italian seismic dataset for machine learning.
\newblock Dataset, 2021.

\bibitem{mousavi2019stead}
S. Mostafa Mousavi, Yixiao Sheng, Weiqiang Zhu, and Gregory C. Beroza.
\newblock Stanford earthquake dataset (STEAD): A global data set of seismic signals for AI.
\newblock IEEE Access, 7:179464--179476, 2019.
\newblock doi: 10.1109/ACCESS.2019.2947848.

\bibitem{mousavi2020eqtransformer}
S. Mostafa Mousavi, William L. Ellsworth, Weiqiang Zhu, Lindsay Y. Chuang, and Gregory C. Beroza.
\newblock Earthquake transformer---an attentive deep-learning model for simultaneous earthquake detection and phase picking.
\newblock Nature Communications, 11, 2020.
\newblock doi: 10.1038/s41467-020-17591-w.

\bibitem{munchmeyer2022whichpicker}
Jannes M{\"u}nchmeyer, Jack Woollam, Andreas Rietbrock, Frederik Tilmann, Dietrich Lange, Thomas Bornstein, Tobias Diehl, Carlo Giunchi, Florian Haslinger, Dario Jozinovi{\'c}, Alberto Michelini, Joachim Saul, and Hugo Soto.
\newblock Which picker fits my data? A quantitative evaluation of deep learning based seismic pickers.
\newblock Journal of Geophysical Research: Solid Earth, 127(1):e2021JB023499, 2022.
\newblock doi: 10.1029/2021JB023499.

\bibitem{puentehuerta2025associators}
Jorge Puente Huerta, Christian Sippl, Jannes M{\"u}nchmeyer, and Ian W. McBrearty.
\newblock Benchmarking seismic phase associators: Method comparison and recommendations for automatic event building.
\newblock Seismica, 4(2), 2025.
\newblock doi: 10.26443/seismica.v4i2.1559.

\bibitem{ross2019phaselink}
Zachary E. Ross, Yisong Yue, Men-Andrin Meier, Egill Hauksson, and Thomas H. Heaton.
\newblock PhaseLink: A deep learning approach to seismic phase association.
\newblock Journal of Geophysical Research: Solid Earth, 124(1):856--869, 2019.
\newblock doi: 10.1029/2018JB016674.

\bibitem{saad2026utrans}
Omar M. Saad, Yangkang Chen, and Tariq Alkhalifah.
\newblock U-Trans: a foundation model for seismic waveform representation and enhanced downstream earthquake tasks.
\newblock Scientific Reports, 2026.
\newblock doi: 10.1038/s41598-026-41454-x.

\bibitem{singh2024conformaleo}
Geethen Singh, Glenn Moncrieff, Zander Venter, Kerry Cawse-Nicholson, Jasper Slingsby, and Tamara B. Robinson.
\newblock Conformal predictions for trustworthy detection of out-of-distribution data in earth observation.
\newblock Scientific Reports, 14:16166, 2024.
\newblock doi: 10.1038/s41598-024-65954-w.

\bibitem{tan2024pavlof}
Darren Tan, David Fee, Alex Witsil, Tarsilo Girona, Matthew Haney, Aaron Wech, Christopher F. Waythomas, and Taryn Lopez.
\newblock Detecting volcanic seismicity during repose at Pavlof volcano, Alaska, using deep learning.
\newblock Journal of Geophysical Research: Solid Earth, 129(6):e2024JB029194, 2024.
\newblock doi: 10.1029/2024JB029194.

\bibitem{myren2025seismicuq}
Samuel Thomas Wilkins Myren, Nidhi K. Parikh, Rosalyn Cherie Rael, Garrison Stevens Flynn, David Mitchell Higdon, and Emily Michele Casleton.
\newblock Evaluation of seismic artificial intelligence with uncertainty.
\newblock Seismological Research Letters, 2025.
\newblock doi: 10.1785/0220240444.

\bibitem{woollam2022seisbench}
Jack Woollam, Jannes M{\"u}nchmeyer, Frederik Tilmann, Andreas Rietbrock, Dietrich Lange, Thomas Bornstein, Tobias Diehl, Carlo Giunchi, Florian Haslinger, Dario Jozinovi{\'c}, Alberto Michelini, Joachim Saul, and Hugo Soto.
\newblock SeisBench---a toolbox for machine learning in seismology.
\newblock Seismological Research Letters, 93(3):1695--1709, 2022.
\newblock doi: 10.1785/0220210324.

\bibitem{zhong2024volcanophase}
Yiyuan Zhong and Y. J. Tan.
\newblock Deep-learning-based phase picking for volcano-tectonic and long-period earthquakes.
\newblock Geophysical Research Letters, 51(12):e2024GL108438, 2024.
\newblock doi: 10.1029/2024GL108438.

\bibitem{zhu2018phasenet}
Weiqiang Zhu and Gregory C. Beroza.
\newblock PhaseNet: A deep-neural-network-based seismic arrival time picking method.
\newblock arXiv preprint arXiv:1803.03211, 2018.

\bibitem{zhu2019deepdenoiser}
Weiqiang Zhu, S. Mostafa Mousavi, and Gregory C. Beroza.
\newblock Seismic signal denoising and decomposition using deep neural networks.
\newblock IEEE Transactions on Geoscience and Remote Sensing, 57:9476--9488, 2019.
\newblock doi: 10.1109/TGRS.2019.2926772.

\bibitem{zhu2022gamma}
Weiqiang Zhu, Ian W. McBrearty, S. Mostafa Mousavi, William L. Ellsworth, and Gregory C. Beroza.
\newblock Earthquake phase association using a bayesian gaussian mixture model.
\newblock Journal of Geophysical Research: Solid Earth, 2022.
\newblock doi: 10.1029/2021JB023249.

\bibitem{zhu2023phasenetdas}
Weiqiang Zhu, Ettore Biondi, Jiaxuan Li, Jiuxun Yin, Zachary E. Ross, and Zhongwen Zhan.
\newblock Seismic arrival-time picking on distributed acoustic sensing data using semi-supervised learning.
\newblock Nature Communications, 14(1):8192, 2023.
\newblock doi: 10.1038/s41467-023-43355-3.

\end{thebibliography}
\end{document}